\documentclass[conference]{IEEEtran}
\IEEEoverridecommandlockouts
\usepackage{cite}
\usepackage{amsmath,amssymb,amsfonts}
\usepackage{algorithmic}
\usepackage{graphicx}
\usepackage{textcomp}
\usepackage{xcolor}
\usepackage{booktabs}
\usepackage{makecell}
\usepackage{threeparttable}
\usepackage{subcaption}
\usepackage{colortbl,xcolor}
\usepackage{authblk}
\usepackage{hyperref}
\usepackage[capitalize]{cleveref}
\crefname{section}{Sec.}{Secs.}
\Crefname{section}{Section}{Sections}
\Crefname{table}{Table}{Tables}
\crefname{table}{Tab.}{Tabs.}

\usepackage{graphicx}
\usepackage{booktabs}
\usepackage{multirow}
\usepackage{rotating}

\def\ie{\textit{i.e.}}

\def\BibTeX{{\rm B\kern-.05em{\sc i\kern-.025em b}\kern-.08em
    T\kern-.1667em\lower.7ex\hbox{E}\kern-.125emX}}

\begin{document}

\title{
PB-UAP: HYBRID UNIVERSAL ADVERSARIAL ATTACK FOR IMAGE SEGMENTATION
}
\author{Yufei Song$^{1}$, Ziqi Zhou$^{2}$, Minghui Li$^{3}$, Xianlong Wang$^{1}$, \\ Hangtao Zhang$^{1}$, 
Menghao Deng$^{1}$, Wei Wan$^{1}$, Shengshan Hu$^{1}$, Leo Yu Zhang$^{4}$} 
\affil{$^{1}$School of Cyber Science and Engineering, Huazhong University of Science and Technology, \\ 
$^{2}$School of Computer Science and Technology, 
Huazhong University of Science and Technology, \\ 
$^{3}$School of Software Engineering, Huazhong University of Science and Technology, \\ 
$^{4}$  School of Information and Communication Technology, Griffith University}

\maketitle
\begin{abstract}
With the rapid advancement of deep learning, the model robustness has become a significant research hotspot, \ie, adversarial attacks on deep neural networks. Existing works primarily focus on image classification tasks, aiming to alter the model's predicted labels. 
Due to the output complexity and deeper network architectures, research on adversarial examples for segmentation models is still limited, particularly for universal adversarial perturbations.
In this paper, we propose a novel universal adversarial attack method designed for segmentation models, which includes dual feature separation and low-frequency scattering modules.
The two modules guide the training of adversarial examples in the pixel and frequency space, respectively.
Experiments demonstrate that our method achieves high attack success rates surpassing the state-of-the-art methods, and exhibits strong transferability across different models.
\end{abstract}

\section{Introduction}
With the development of deep learning, 
semantic segmentation models are playing an increasingly important role in complex scenarios such as autonomous driving~\cite{xiao2023baseg}, medical image analysis~\cite{kalinin2020medical}, and remote sensing~\cite{du2021incorporating}.
By classifying semantic information at the pixel level, segmentation models~\cite{zhao2017pyramid,chen2014semantic,chen2017deeplab} achieve accurate object segmentation.
However, recent works~\cite{gu2022segpgd,jia2023transegpgd} demonstrate that segmentation models are vulnerable to adversarial attacks, 
where imperceptible noise is added to images, leading to incorrect model predictions.



Existing adversarial attacks can be categorized into sample-wise~\cite{li2024transfer,zhou2025numbod} and universal adversarial perturbations (UAPs)~\cite{machado2021adversarial,zhou2023downstream, zhou2023advclip,zhou2024darksam}, where UAP refers to a single perturbation applied to various examples, causing the model to produce erroneous outputs across different inputs. 
Despite the promising attack performance of UAPs in classification tasks, the UAP methods for segmentation models~\cite{hendrik2017universal,hashemi2022improving} still fall short. The earliest UAP method~\cite{hendrik2017universal} for segmentation builds on UAP~\cite{moosavi2017universal} by averaging gradients over each batch to mislead the segmentation model. Another study~\cite{hashemi2022improving} enhances the generalization of attacks across different segmentation models by utilizing feature similarities of input examples in the first layer of the model, based on the UAPGD~\cite{deng2020universal} method. Although these methods optimize the traditional UAP approach, they still do not achieve satisfactory attack performance. We attribute this limitation to the direct adaptation of classification-oriented UAP techniques, which overlooks task-specific knowledge critical to segmentation.
%
%
%
%
 \begin{figure}[t]
    \centering
    \includegraphics[scale=0.32]{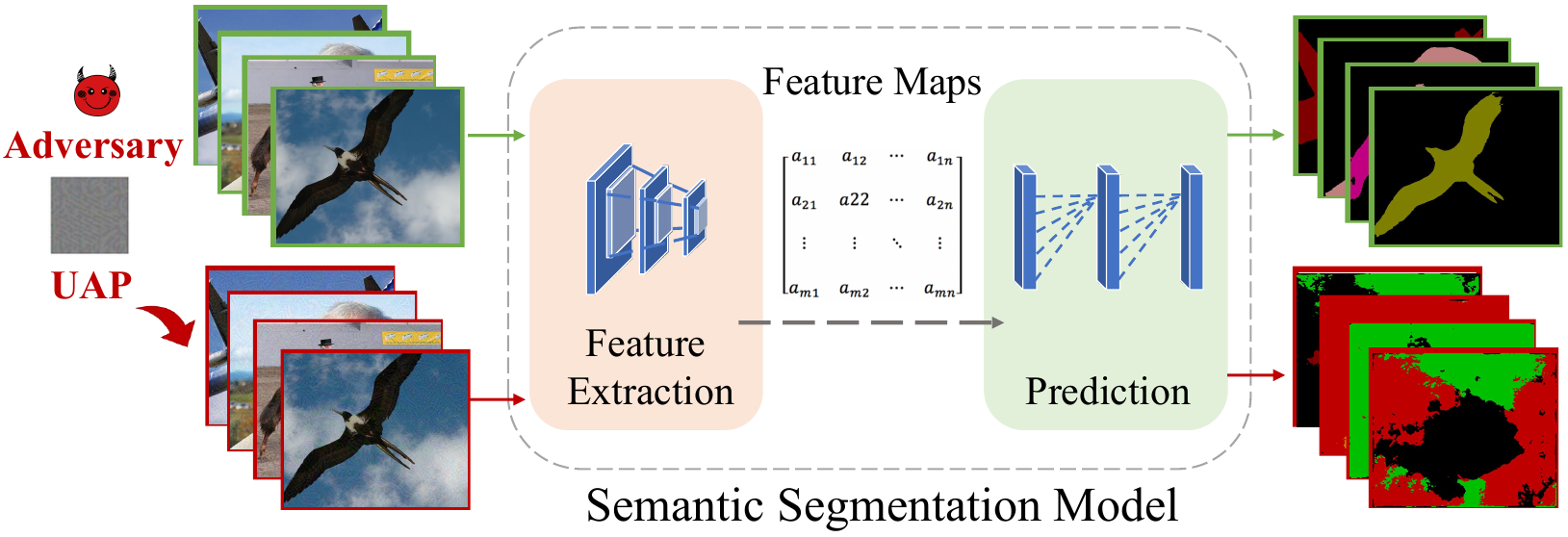}
    \caption{
    Illustration of fooling models using a UAP.
    }
    \label{fig:demo}
  \vspace{-0.4cm}
\end{figure}

In this paper, we propose \underline{\textbf{P}}ixel \underline{\textbf{B}}lind \underline{\textbf{UAP}} (PB-UAP), a novel universal adversarial attack in segmentation tasks that disrupts image features in both spatial and frequency domains. Our method aims to disable the models segmentation ability across diverse images using a single UAP (see \cref{fig:demo}). Unlike classification models that focus on global features, segmentation models concentrate more on contextual relationships within images~\cite{chen2014semantic,zhao2017pyramid,gao2023prototype}.
Therefore, our intuition is to destroy inter-class and intra-class semantic correlations in the image to mislead the model into incorrectly segmenting the input images. 
In the spatial domain, we deviate the output features of adversarial examples from both the output features of benign examples and the ground truth labels, aiming to undermine inter-class semantic correlations.
In the frequency domain, given that pixels within the same class primarily belong to the low-frequency components of images, we separate the low-frequency features between adversarial and original examples to disrupt intra-class semantic correlations.

In conclusion, our main contributions are three-folds. \textbf{1)} We propose an effective universal adversarial attack for segmentation tasks, dubbed PB-UAP, disrupting image features in both the spatial and frequency domains. \textbf{2)} We propose a dual feature separation and low-frequency scattering strategy that overcomes the limitations of inter-class and intra-class semantic correlations. \textbf{3)} Experimental results on three models and two benchmark datasets demonstrate that PB-UAP significantly outperforms state-of-the-art methods, and exhibits strong transferability across different models.

\section{Background and Related Works}
\subsection{Semantic Segmentation Models}
Segmentation models typically incorporate architectures such as Image Pyramid~\cite{wu2022fpanet}, Encoder-Decoder~\cite{xing2020encoder}, Context Module~\cite{wu2020cgnet}, Spatial Pyramid Pooling~\cite{ru2023forest}, Atrous Convolution (AC)~\cite{chen2014semantic}, and Atrous Spatial Pyramid Pooling (ASPP)~\cite{chen2017deeplab}.
These architectures enhance the model's ability to capture image details by introducing multi-scale feature extraction, strengthening contextual information, or expanding the receptive field, resulting in more precise segmentation outcomes.
Specifically, Deeplabv1~\cite{chen2014semantic} and PSPNet~\cite{zhao2017pyramid} uses AC structure to enhance global context awareness while preserving local details without sacrificing resolution.
After Deeplabv2~\cite{chen2017deeplab}, subsequent versions~\cite{chen2017rethinking,chen2018encoder} adopted the ASPP structure, which enhances the model's understanding of complex scenes and object boundaries through multi-scale feature aggregation. 
Different models have distinct architectures, resulting in varying vulnerabilities to adversarial attacks.


\subsection{Universal Adversarial Attacks for Segmentation}
Deep learning models are vulnerable to poisoning attacks~\cite{zhangdenial,wangunlearnable}, backdoor attacks~\cite{hu2022badhash,zhang2024detector,wang2024trojanrobot,yao2024reverse}, and adversarial attacks~\cite{zhang2024badrobot,yichen}, among which standard universal adversarial attack methods~\cite{moosavi2017universal,deng2020universal}
show outstanding performance in classification tasks. 
However, when applied to segmentation tasks, they exhibit limitations in disrupting the model's understanding of contextual information.
The first UAP method~\cite{hendrik2017universal} for segmentation extends UAP~\cite{moosavi2017universal} by averaging gradients over each batch, enabling targeted attacks on street scenes. However, this approach neglects the semantic interdependencies between classes, limiting the generalization of adversarial examples. 
To address this, a subsequent work~\cite{zhang2021data} enhances generalization by applying data augmentation techniques that inject high-frequency information into the training images. 
Another research~\cite{hashemi2022improving} demonstrate that similar feature representations in the initial layers of different models improve cross-model attack transferability. Despite these advancements, the attack success rate remains hindered by intra-class semantic dependencies.
Therefore, existing work does not address the limitations of classification-based UAP methods in segmentation tasks, indicating the need for a more effective strategy.

\section{Methodology}

 \begin{figure*}[!t]
    \centering
    \includegraphics[scale=0.58]{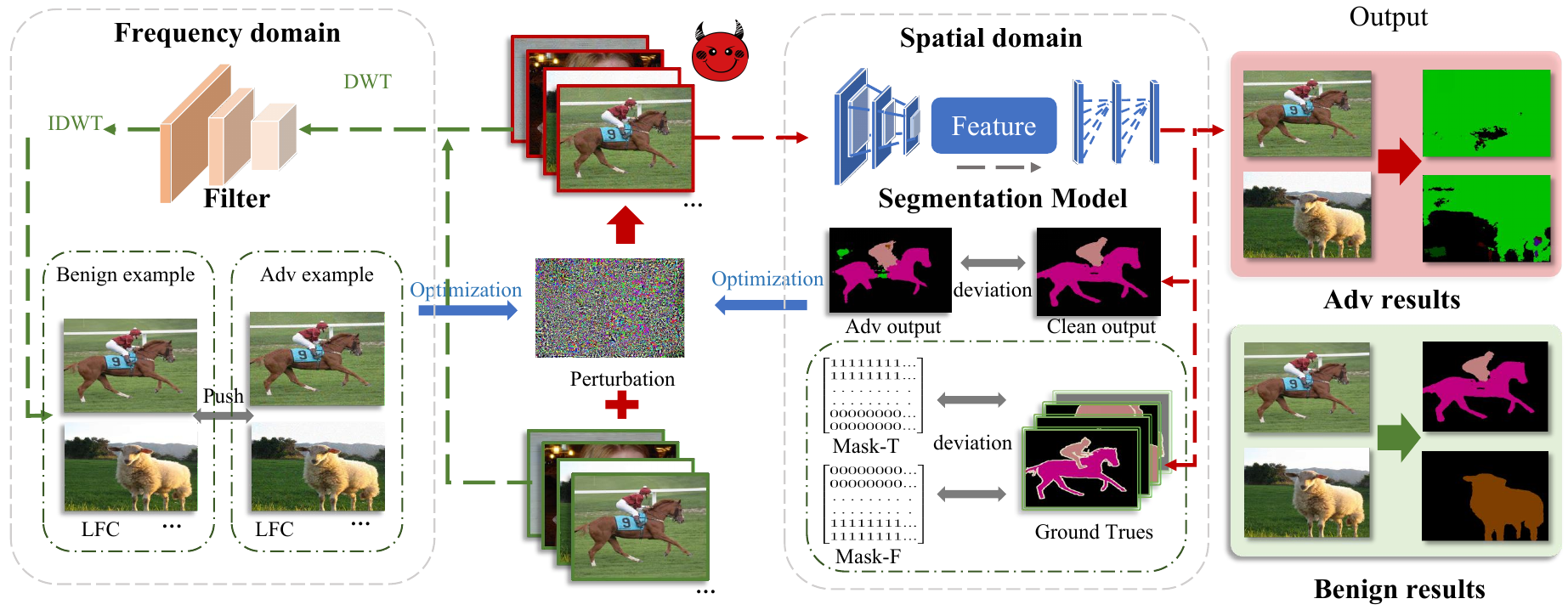}
    \caption{The framework of PB-UAP.
    }
    \label{fig:pipeline}
      \vspace{-0.2cm}
\end{figure*}

\subsection{Problem Formulation}

In semantic segmentation, adversarial attacks can be defined as adding imperceptible perturbations to deceive the model, causing incorrect classifications for every pixel in the image. 
Specifically, let \( f(x) \) denote the segmentation model, where \( x \in \mathbb{R}^{W \times H \times C} \) represents the input image, and the model’s output consists of the predicted class labels for each pixel.
This optimization problem can be formulated as maximizing the cross entropy loss function $L_{CE}$ by adjusting the perturbation $\delta$ within the constraint set $S$, thereby increasing the discrepancy between the predictions and the ground truth labels \( y^{\text{true}} \):
\begin{equation}
\delta^* = \arg \max_{\delta \in S} \sum L_{CE}(f(x\ + \delta), y^{\text{true}}).
\label{eq:jt}
\end{equation}

\subsection{Intuition Behind PB-UAP}
A successful universal adversarial attack on segmentation task should apply a UAP to induce incorrect predictions for all pixels across different examples, requiring the attack to consider both the global context and local details of the images. Specifically, it faces the following two challenges:

\noindent\textbf{Challenge I: Inter-class Semantic Correlations Limit the Universality of Attacks.}
Segmentation tasks involve complex semantic correlations, including class boundaries, object shapes, and structures, making the optimization of perturbations more challenging. A universal adversarial attack must generate a single perturbation that generalizes across diverse examples, addressing variations in class boundaries, shapes, and structures. This requires identifying a common vulnerability in the model. 
To this end, we propose a dual feature deviation strategy, applying gradient ascent on the features of the model's final layer and the pixel-level labels. The former disrupts the model's global semantic understanding, weakening its reliance on class boundaries, while the latter ensures the perturbation affects diverse target labels, bypassing semantic constraints related to shapes and structures.

\noindent\textbf{Challenge II: Intra-class Semantic Correlations Weaken the Attacks.}
Intra-class semantic correlations refer to the spatial relationships between pixels of the same target in an image. This correlation is evident in adjacent pixels having similar visual features and semantic information. Specifically, these pixels form coherent and similar local features in the image. When adversarial perturbations affect these regions, the model smooths the impact of the perturbations by understanding the local context. This contextual integration makes it difficult for perturbations to have significant effects within these regions, thereby reducing the success rate of adversarial attacks. Considering that in the frequency domain analysis of an image, the low-frequency components contain the overall structure and most of the smooth information of the image, while the high-frequency components, such as the edges and textures of objects, represent less semantic information, and perturbations in these areas are influenced by local context smoothing. Based on this consideration, we use low-frequency scattering to disrupt pixel correlations. Specifically, we separate the low-frequency components of adversarial examples from the original examples and increase the perturbation strength in the low-frequency regions to improve the effectiveness of the attack.

\subsection{PB-UAP: A Complete Illustration}

In this section, we introduce PB-UAP, a hybrid spatial-frequency universal adversarial attack method designed for semantic segmentation tasks. The pipeline of PB-UAP is illustrated in~\cref{fig:pipeline}, including a spatial attack based on dual feature deviation and a frequency attack based on low-frequency scattering. Specifically, we separate the output features of adversarial examples and benign examples at the final layer of the model, while separating the outputs of adversarial examples from the ground truth labels, in order to disrupt the semantic correlation between different target classes. Additionally, we separate the low-frequency components of adversarial examples and clean examples to break the semantic correlation between adjacent pixels of the same class, further enhancing the attack success rate.  The overall optimization objective can be summarized as:
\begin{equation}
\mathcal{J}_{\text{total}}=\mathcal{J}_{pd}+\mathcal{J}_{fd} + k*\mathcal{J}_{ls}.
\label{eq:jt}
\end{equation}
where $\mathcal{J}_{pd}$ and $\mathcal{J}_{fd}$ denote the pixel-level deception attack loss and the feature distortion attack loss, respectively, while $\mathcal{J}_{ls}$ represents the low-frequency scattering loss.

\noindent\textbf{Dual Feature Deviation} 
In this module, we define two loss functions as optimization objectives, including the deviation loss between the output features of adversarial examples and both the output features of original examples and the ground truth labels.
First, Eq. (\ref{eq:success}) and Eq. (\ref{eq:fail}) calculate the difference between output features of the adversarial examples and the ground truth labels. 
$\delta$ denotes the perturbation, $M$ is a binary matrix, where zeros and ones represent the pixel locations in the segmentation mask that are misclassified and correctly classified, respectively. 
$f(x + \delta)$ and $y_{true}$ are the outputs of the adversarial example and the ground truth labels, respectively. $ J_{\text{suc}} $ and $ J_{\text{fail}} $ represent the loss for pixels in which the attack succeeds and fails, respectively. 
\begin{equation}
    \mathcal{J}_{\text{suc}} = -(L_{CE}(f(x + \delta), y_{true})) * M,
    \label{eq:success}
\end{equation}
\begin{equation}
    \mathcal{J}_{\text{fail}} = -(L_{CE}(f(x + \delta), y_{true})) * \overline{M},
    \label{eq:fail}
\end{equation}
Eq. (\ref{eq:label_loss}) illustrates the process of assigning different optimization weights to correctly and incorrectly classified pixels. 
The hyperparameter $\lambda$, with a value of 0.3, represents the weight assigned to correctly classified pixels, while 1 - $\lambda$ represents the weight assigned to incorrectly classified pixels.
\begin{equation}
    \mathcal{J}_{pd} = \lambda \cdot \mathcal{J}_{\text{suc}} + (1 - \lambda) \cdot \mathcal{J}_{\text{fail}},
    \label{eq:label_loss}
\end{equation}
Secondly, Eq. (\ref{eq:feature_loss}) calculates the difference between the outputs of adversarial examples and benign examples, where $L_{MSE}$ is the mean-square error loss function. 
\begin{equation}
\mathcal{J}_{fd} = -(L_{MSE}(f(x + \delta), f(x))).
\label{eq:feature_loss}
\end{equation}

\noindent\textbf{Low-frequency Scattering } 
In segmentation tasks, the low-frequency regions of an image contain most of the semantic information for each object, where semantic consistency between adjacent pixels exhibits a strong spatial correlation. 
This spatial correlation smooths the effect of perturbations, thereby reducing attack performance. 
To disrupt this correlation, we use the discrete wavelet transform (DWT)~\cite{luo2022frequency} with a low-pass filter $\mathcal{L}$ to decompose the image into its low-frequency component $c_{ll}$.
Next, we use the inverse discrete wavelet transform (IDWT) to reconstruct the low-frequency component into an image $\phi (x) $, as detailed in Eq.~\ref{eq:dwt}. 
Subsequently, we perform low-frequency scattering by calculating the mean square error between the low-frequency images of the adversarial and original examples, thereby disrupting the correlation between adjacent pixels, as shown in Eq.~\ref{eq:low}.
\begin{equation}
  \label{eq:dwt}
  c_{ll}   =  \mathcal{L} x  \mathcal{L}^T, \phi (x)   =  \mathcal{L} ^T c_{ll}\mathcal{L}  = \mathcal{L} ^T (\mathcal{L}  x \mathcal{L}^T)\mathcal{L},
\end{equation}
\begin{equation} 
\label{eq:low}
\mathcal{J}_{ls} =-(L_{MSE}\left ( \phi (x + \delta),\phi (x\right )). 
\vspace{0.1cm}
\end{equation}

\section{Experiments}

\subsection{Experimental Setup}

\noindent\textbf{Datasets and Models.} 
We use two public segmentation datasets to evaluate the attack performance of our method: PASCAL VOC 2012~\cite{everingham2010pascal} and CITYSCAPES~\cite{cordts2016cityscapes}. 
We choose PSPNet, Deeplabv1, and Deeplabv3+ with MobileNet and ResNet50 backbones as victim models. 

\noindent\textbf{Parameter Setting.} 
Following~\cite{moosavi2017universal,deng2020universal,zhou2024securely}, we set the upper bound of UAP to $10/255$.
For our experiments, we set the hyperparameters $k$, $\lambda$, and the batch size to $1$, $0.3$, and $5$, respectively. 

\noindent\textbf{Evaluation Metrics.} 
We use the Mean Intersection over Union (mIoU) to evaluate the effectiveness of PB-UAP, which is a popular metric in the segmentation field. 
A lower mIoU indicates stronger attack effectiveness.
\subsection{Attack Performance}
\begin{table}[htbp]
  \centering
  \caption{The mIoU (\%) of PB-UAP under different settings. Values covered by gray denote the mIOU of benign examples, others denote the mIOU of adversarial examples.}
  \resizebox{0.49\textwidth}{!}{%
    \begin{tabular}{ccccccc}
    \toprule
    \toprule
    \multirow{2}[4]{*}{Dataset} & \multicolumn{2}{c}{PSPNet} & \multicolumn{2}{c}{Deeplabv1} & \multicolumn{2}{c}{Deeplabv3+} \\
\cmidrule{2-7}          & MobileNet & ResNet50 & MobileNet & ResNet50 & MobileNet & ResNet50 \\
    \midrule
    \multirow{2}[1]{*}{VOC} & \cellcolor[rgb]{ .949,  .949,  .949}66.65 & \cellcolor[rgb]{ .949,  .949,  .949}71.58 & \cellcolor[rgb]{ .949,  .949,  .949}53.5 & \cellcolor[rgb]{ .949,  .949,  .949}58.9 & \cellcolor[rgb]{ .949,  .949,  .949}70.62 & \cellcolor[rgb]{ .949,  .949,  .949}71.37 \\
          & 10.48 & 16.54 & 9.73  & 14.25 & 12.19 & 18.77 \\
    \multirow{2}[1]{*}{CITYSCAPES} & \cellcolor[rgb]{ .949,  .949,  .949}56.91 & \cellcolor[rgb]{ .949,  .949,  .949}60.58 & \cellcolor[rgb]{ .949,  .949,  .949}56.49 & \cellcolor[rgb]{ .949,  .949,  .949}60.27 & \cellcolor[rgb]{ .949,  .949,  .949}61.94 & \cellcolor[rgb]{ .949,  .949,  .949}64.67 \\
          & 6.02  & 8.6   & 6.61  & 8.71  & 3.17  & 4.41 \\
    \bottomrule
    \bottomrule
    \end{tabular}%
    }
  \label{tab:main}%
\end{table}%

In this section, we comprehensively evaluate the effectiveness of PB-UAP. We conduct attack experiments on three segmentation models and two datasets, with two backbone networks. We calculate the mIoU of both benign and adversarial examples for each experimental setup.

According to the experimental results in~\cref{tab:main}, the mIoU dropped to between 3.17\% and 18.77\% in all experimental setups, indicating that PB-UAP can effectively influence the output of segmentation models. 
\subsection{Transferability Study}
\begin{figure}[!t]
    \centering
    \subcaptionbox{PASCAL VOC}[0.5\linewidth][c]{
        \includegraphics[scale=0.32]{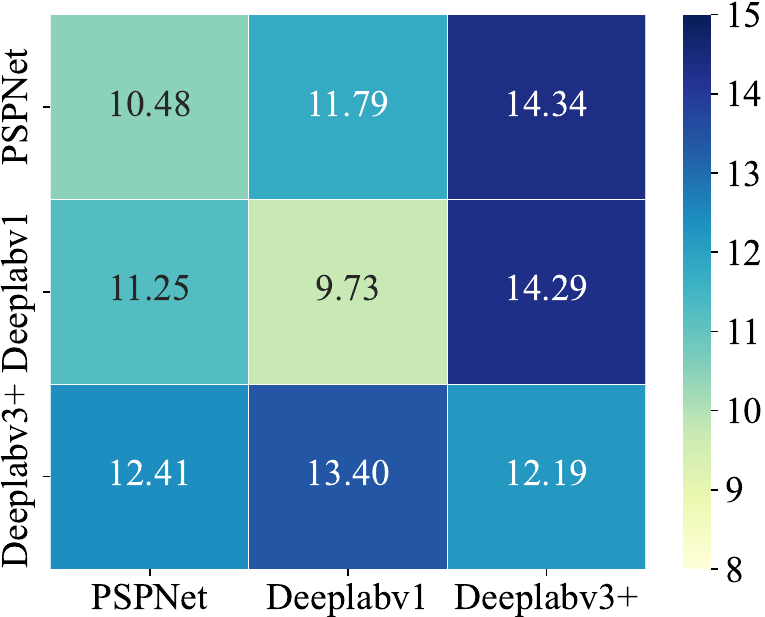}
    }\hfill
    \subcaptionbox{CITYSCAPES}[0.5\linewidth][c]{
        \includegraphics[scale=0.32]{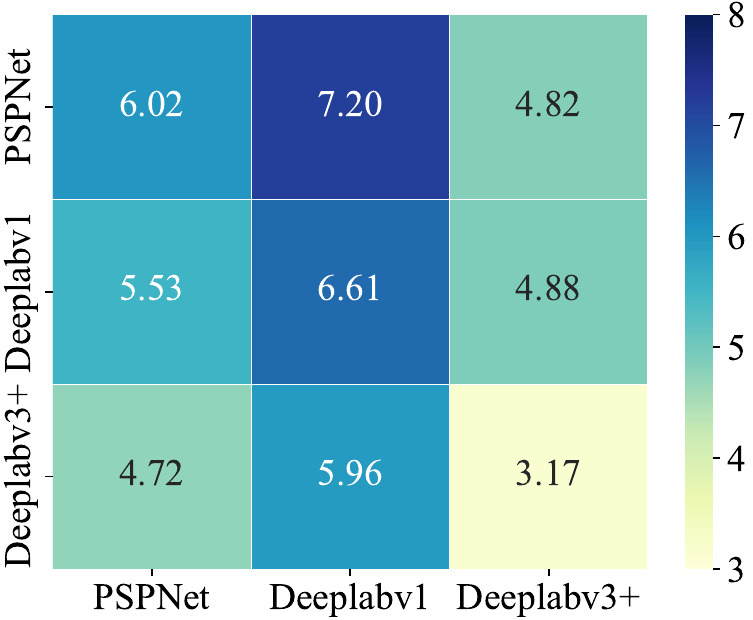}
    }
    
    \caption{Transferability study. Each column represents attacking different modles using the same adversarial examples.}
    \label{fig:transferability}
    \vspace{-0.5cm}
\end{figure}
We investigate the attack transferability of PB-UAP across different models, using MobileNet as the backbone network. As shown in~\cref{fig:transferability}(a) - (b), the UAPs generated using different proxy models exhibit excellent performance on other models. These results demonstrate that PB-UAP possesses strong transferability and reliability.
\subsection{Comparison Study}
%
\begin{table}[htbp]
  \centering
  \caption{The mIoU(\%) result of comparison study.}
  \resizebox{0.49\textwidth}{!}{%
    \begin{tabular}{ccccccc}
    \toprule
    \toprule
    \multirow{2}[4]{*}{Method} & \multicolumn{2}{c}{PSPNet} & \multicolumn{2}{c}{Deeplabv1} & \multicolumn{2}{c}{Deeplabv3+} \\
\cmidrule{2-7}          & MobileNet & ResNet50 & MobileNet & ResNet50 & MobileNet & ResNet50 \\
    \midrule
    Benign & 66.65  & 71.58  & 53.50  & 58.90  & 70.62  & 71.37  \\
    UAPGD & 45.37  & 57.26  & 28.81  & 40.68  & 57.38  & 57.95  \\
    FFF   & 40.36  & 49.98  & 53.49  & 58.90  & 48.74  & 51.47  \\
    Hashemi   & 34.19  & 44.20  & 25.97  & 35.17  & 49.50  & 49.84  \\
    SegPGD & 34.82  & 44.87  & 25.30  & 34.58  & 47.48  & 49.89  \\
    TranSegPGD & 33.30  & 47.77  & 24.35  & 33.96  & 40.73  & 50.02  \\
    Ours  & \textbf{10.48 } & \textbf{16.54 } & \textbf{9.73 } & \textbf{14.25 } & \textbf{12.19 } & \textbf{18.77 } \\
    \bottomrule
    \bottomrule
    \end{tabular}%
    }
  \label{tab:addlabel}%
\end{table}%
%
%
To demonstrate the superiority of our method, we compare PB-UAP with five previous popular UAP schemes, including SegPGD~\cite{gu2022segpgd} and TranSegPGD~\cite{jia2023transegpgd}, which are state-of-the-art adversarial attacks designed for segmentation models. We select PASCAL VOC~\cite{everingham2010pascal} as the attacked dataset, with model and backbone settings as described in Section 4.2.
 \begin{figure}[t]
  \setlength{\abovecaptionskip}{4pt}
  \setlength{\belowcaptionskip}{-1em}
    \centering
    \includegraphics[scale=0.425]{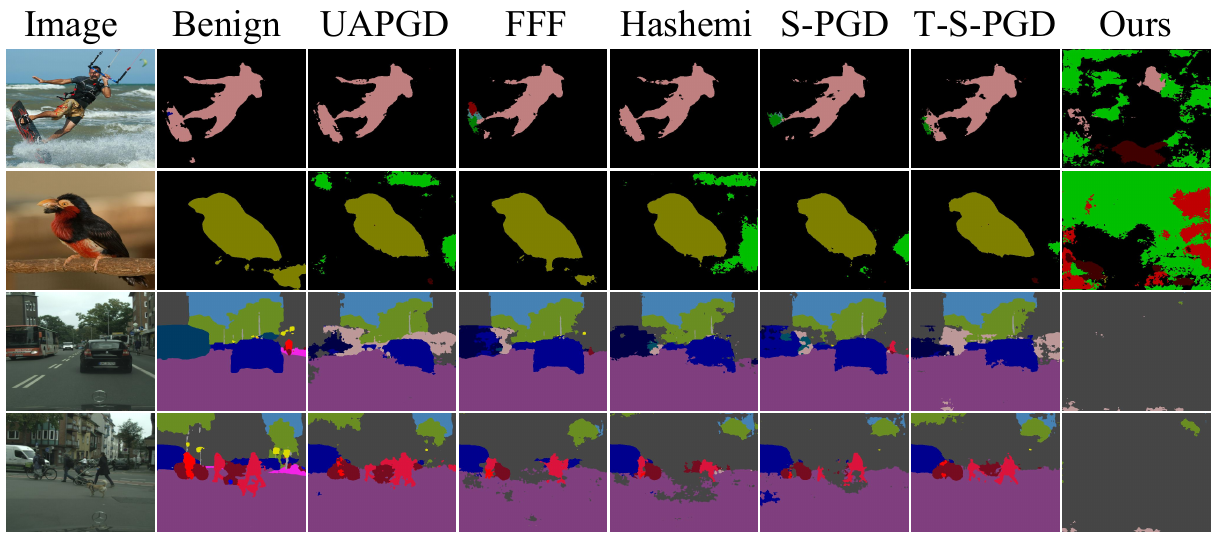}
    \caption{Visualizations of the comparison study.
    }
    \label{fig:compare}
     \vspace{-0.2cm}
\end{figure}

The results in~\cref{tab:addlabel} indicate that PB-UAP outperforms all methods significantly. 
We also provide visualizations of the segmentation results of the adversarial examples generated by these methods in~\cref{fig:compare}, which further demonstrate the superiority of PB-UAP.
\begin{figure}[h]
    \centering
    
    \subcaptionbox{Modules}[0.5\linewidth][c]{
        \includegraphics[scale=0.176]{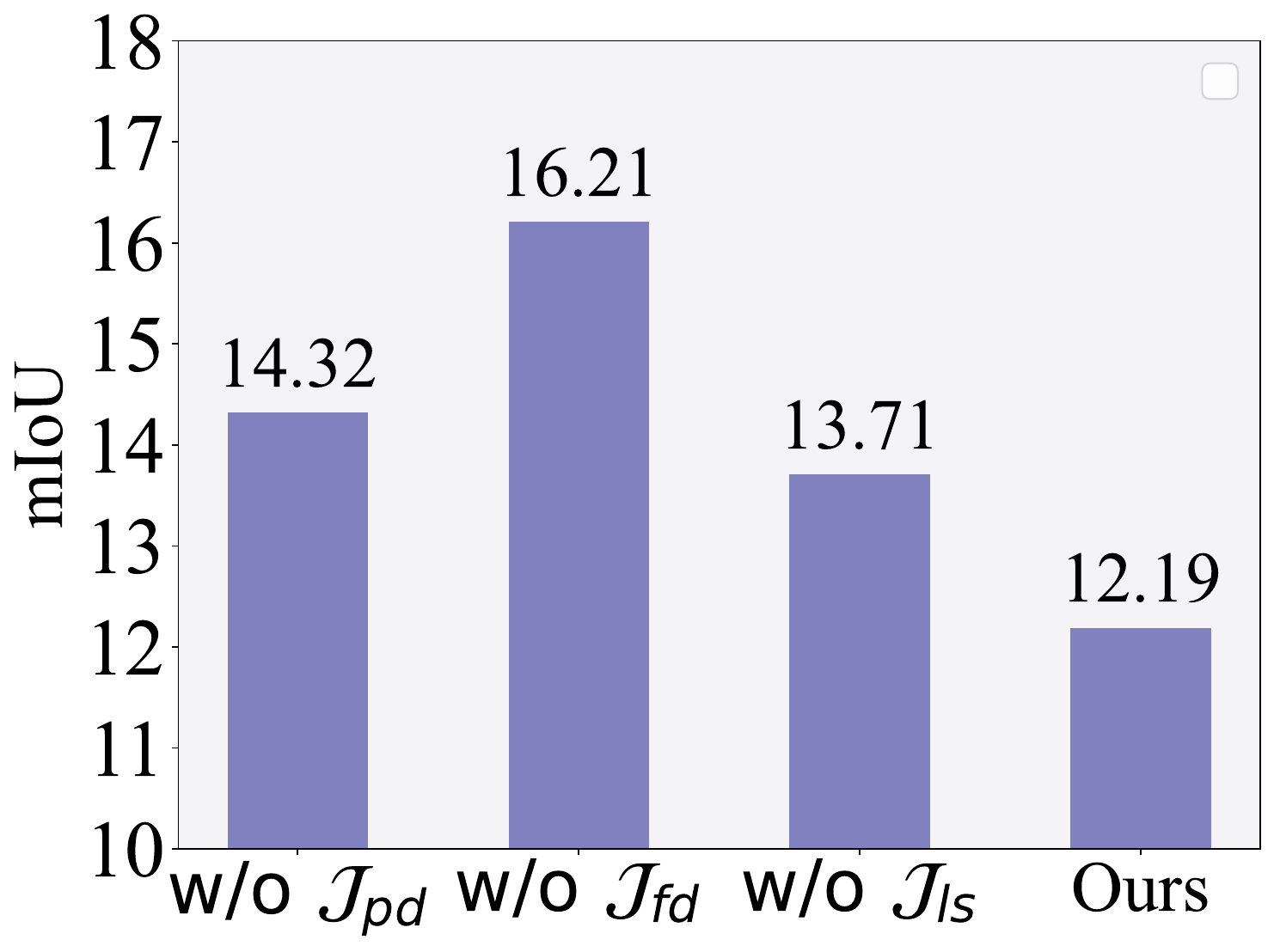}
    }\hfill
    \subcaptionbox{Epsilon}[0.5\linewidth][c]{
        \includegraphics[scale=0.175]{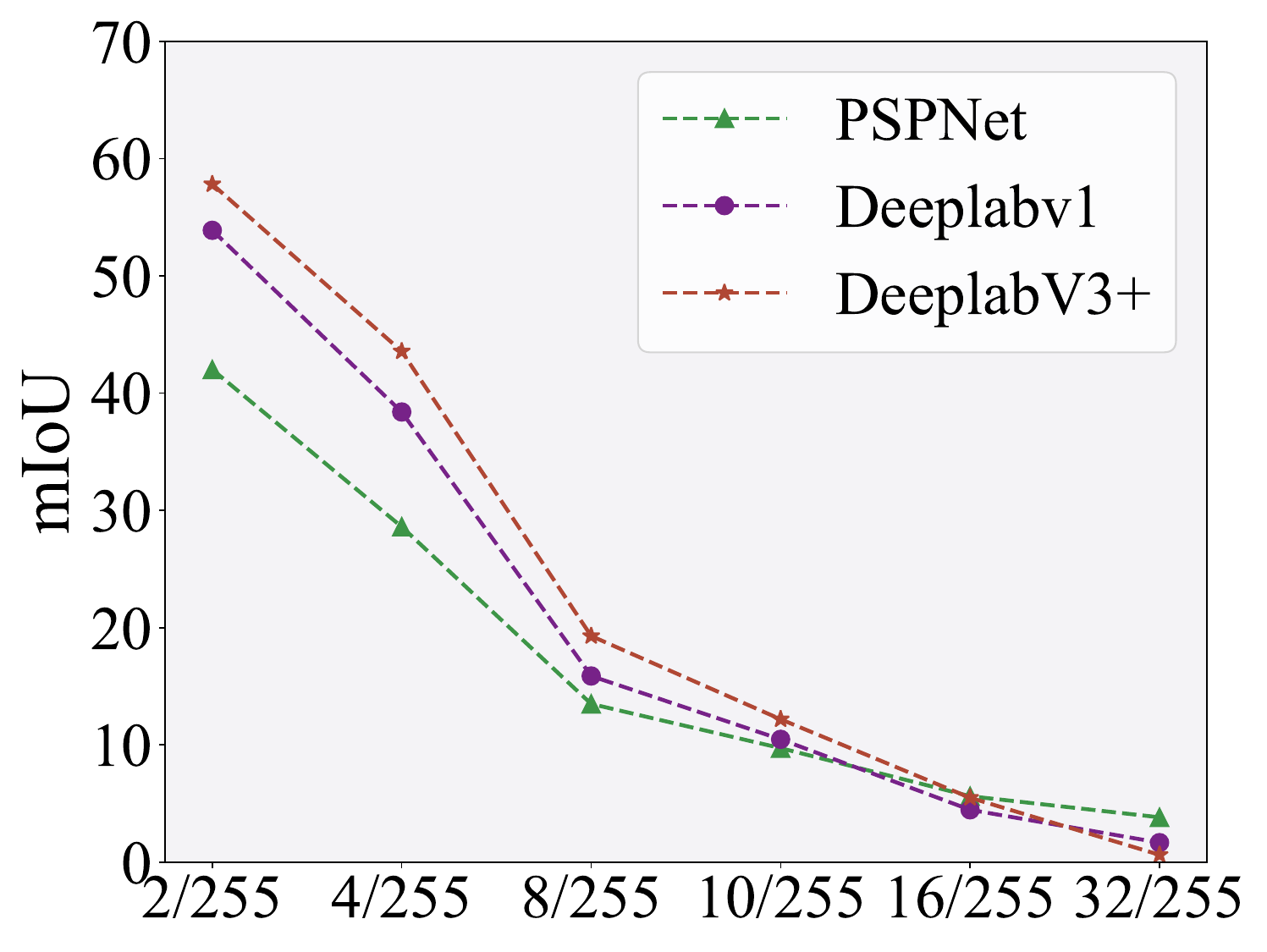}
        \vspace{0.05cm}
    }
    
    \caption{Ablation study results (\%). (a) and (b) investigate the effect of different modules and attack strengths in fake mask on PB-UAP.}
    \label{fig:abaltion_study}
    \vspace{-0.3cm}
\end{figure}
\subsection{Abaltion Study}
\noindent\textbf{The effect of different modules.} 
We investigate the effect of various modules on the attack performance of PB-UAP, with Deeplabv3+ as the model and MobileNet as the backbone network. 
The results in \cref{fig:abaltion_study} (a) show that no variants can compete with the complete method, implying the indispensability of each component for PB-UAP. 

\noindent\textbf{The effect of perturbation budget.} 
As shown in \cref{fig:abaltion_study} (b), we evaluate PB-UAP's attack performance in different values of $\epsilon$, using MobileNet as the backbone network. With the increase in $\epsilon$ , there is a corresponding enhancement in attack performance. 
Notably, our attack still maintains high efficacy at the $8/255$ setting, with an average mIoU exceeding $16.22\%$.

\section{Conclusion}


In this paper, we propose PB-UAP, a universal adversarial
attack specifically designed for the characteristics of segmentation models. 
PB-UAP can effectively induce incorrect segmentation results in the model across different input examples with a single perturbation. To disrupt both inter-class and intra-class semantic correlations in images, we design
a hybrid spatial-frequency universal attack framework. 
This framework consists of a dual feature deviation-based spatial attack and a low-frequency scattering-based frequency attack.
We conduct extensive experiments on PSPNet, Deeplabv1,
and Deeplabv3+ models. 
Both the qualitative and quantitative results demonstrate PB-UAP’s high attack success rates and strong attack transferability.

\section*{Acknowledgments}
This work is supported by the National Natural Science Foundation of China (Grants No. 62202186, No. 62372196). Ziqi Zhou is the corresponding author.

\bibliographystyle{IEEEbib}
\bibliography{ICASSP2024/main}

\end{document}